\documentclass[conference]{IEEEtran}
\IEEEoverridecommandlockouts
\usepackage{cite}
\usepackage{amsmath,amssymb,amsfonts}
\usepackage{algorithmic}
\usepackage{graphicx}
\usepackage{textcomp}
\usepackage{xcolor}

\usepackage{booktabs}
\usepackage{multirow}
\usepackage{lipsum}
\ifCLASSOPTIONcompsoc
\usepackage[caption=false, font=normalsize, labelfont=sf, textfont=sf]{subfig}
\else
\usepackage[caption=false, font=footnotesize]{subfig}
\fi

\usepackage[lined,ruled,linesnumbered,commentsnumbered]{algorithm2e}

\def\BibTeX{{\rm B\kern-.05em{\sc i\kern-.025em b}\kern-.08em
    T\kern-.1667em\lower.7ex\hbox{E}\kern-.125emX}}
\begin{document}

\title{Balancing Exploration and Exploitation for Solving Large-scale Multiobjective Optimization via Attention Mechanism
}

\author{\IEEEauthorblockN{1\textsuperscript{st} Haokai Hong}
\IEEEauthorblockA{\textit{School of Informatics} \\
\textit{Xiamen University}\\
Fujian, China. \\
haokaihong@stu.xmu.edu.cn}
\and
\IEEEauthorblockN{2\textsuperscript{nd} Min Jiang}
\IEEEauthorblockA{\textit{School of Informatics} \\
\textit{Xiamen University}\\
Fujian, China. \\
minjiang@xmu.edu.cn}
\and
\IEEEauthorblockN{3\textsuperscript{rd} Liang Feng}
\IEEEauthorblockA{\textit{College of Computer Science} \\
\textit{Chongqing University}\\
Chongqing, China \\
liangf@cqu.edu.cn}
\and
\IEEEauthorblockN{4\textsuperscript{th} Qiuzhen Lin}
\IEEEauthorblockA{\textit{College of Computer Science and Software Engineering} \\
\textit{Shenzhen University}\\
Guangdong, China \\
qiuzhlin@szu.edu.cn}
\and
\IEEEauthorblockN{5\textsuperscript{th} Kay Chen Tan}
\IEEEauthorblockA{\textit{Department of Computing} \\
\textit{The Hong Kong Polytechnic University}\\
Hong Kong SAR \\
kctan@polyu.edu.hk}
    \thanks{H. Hong and M. Jiang are with the Department of Artificial Intelligence, Key Laboratory of Digital Protection and Intelligent Processing of Intangible Cultural Heritage of Fujian and Taiwan, Ministry of Culture and Tourism, School of Informatics, Xiamen University, Fujian, China, 361005.}
    \thanks{The corresponding author: Min Jiang, minjiang@xmu.edu.cn}
}


\maketitle

\begin{abstract}

Large-scale multiobjective optimization problems (LSMOPs) refer to optimization problems with multiple conflicting optimization objectives and hundreds or even thousands of decision variables. A key point in solving LSMOPs is how to balance exploration and exploitation so that the algorithm can search in a huge decision space efficiently. Large-scale multiobjective evolutionary algorithms consider the balance between exploration and exploitation from the individual's perspective. However, these algorithms ignore the significance of tackling this issue from the perspective of decision variables, which makes the algorithm lack the ability to search from different dimensions and limits the performance of the algorithm. In this paper, we propose a large-scale multiobjective optimization algorithm based on the attention mechanism, called (LMOAM). The attention mechanism will assign a unique weight to each decision variable, and LMOAM will use this weight to strike a balance between exploration and exploitation from the decision variable level. Nine different sets of LSMOP benchmarks are conducted to verify the algorithm proposed in this paper, and the experimental results validate the effectiveness of our design.

\end{abstract}


\section{Introduction}
\par
Optimization problems that involve hundreds or even thousands of decision variables and multiple conflicting objectives are defined as large-scale multiobjective optimization problems (LSMOPs) \cite{tian2021evolutionary}. Critical problems faced in network science\cite{8618599}, machine learning\cite{8482477}, scheduling\cite{RN258}, and biotechnology \cite{TAN2005485} can be solved by solving LSMOPs \cite{hong2021evolutionary}. However, conventional multiobjective evolutionary algorithms (MOEAs) \cite{996017, RN135, 6848830} are severely affected since the decision space grows exponentially with the number of decision variables. This problem is known as the curse of dimensionality of the LSMOP\cite{tian2021evolutionary}, which makes it difficult to apply MOEAs to complex real-world problems.
\par
To address the curse of dimensionality, many large-scale MOEAs have also been developed for solving LSMOPs \cite{RN106, RN89, RN222}. CCGDE3, proposed by Antonio \emph{et al.} \cite{6557903}, is a representative MOEA for solving LSMOPs based on decision variable grouping, which can directly convert an LSMOP into several small-scale multiobjective optimization problems to be solved in sequence. Decision space reduction based methods can quickly find some local optimal solutions for LSMOPs, and the framwork MOEA/PSL proposed by Tian \emph{et al.} \cite{RN85} is typical in this category. The third category is novel search strategy based approaches, GMOEA proposed by He \emph{et al.} \cite{RN90} indicates that novel reproduction operators can effectively solve LSMOPs without the help of decision variable grouping or decision space reduction.
\par
Large-scale MOEAs focus on decision variables and use analysis, grouping, or dimensionality reduction on decision variables, which aim at guiding exploration and exploitation\cite{10.1145/2480741.2480752}. However, the result of a multiobjective optimization problem is a set of solutions with different distributions in different dimensions. And different decision variables have different effects on exploration and exploitation during the optimization process. Therefore, without considering how to balance exploration and exploitation on the decision variable level, it is difficult to further improve the performance of the algorithm.
\par
Based on the above consideration, we design a large-scale multiobjective optimization algorithm focusing on balancing the exploitation and exploration at the decision variable level via the attention mechanism (LMOAM) for solving LSMOPs. Inspired by the application of the attention mechanism in machine learning\cite{10.5555/3295222.3295349}, the proposed algorithm calculates attention for different decision variables so that each decision variable will be guided to explore or exploit by attention. Specifically, the algorithm analyzes the variance of different decision variables and then groups the decision variables accordingly. Afterward, the algorithm calculates and searches attention to each decision variable. Finally, different decision variables perform different exploration or exploitation according to the assigned attention. The main new contributions are summarized as follows.

\begin{enumerate}
    \item A decision variable analysis strategy based on the attention mechanism is suggested to determine the different roles of different decision variables during the optimization. Therefore the algorithm can balance exploration and exploitation at the level of decision variables.
    \item On this basis, a large-scale optimization algorithm based on the attention mechanism is proposed, which improves the algorithm's performance to solve the LSMOP by optimizing decision variables with different attention.
\end{enumerate}

The remainder of this paper is organized as follows. In Section \ref{sec:Preliminaries-and-Related}, we briefly recall some preliminary studies and discuss some related works. The details of our proposed LMOAM method are presented in Section \ref{sec:LMOAM}. In Section \ref{sec:exp}, we conduct a series of experiments compared with some state-of-the-art large-scale MOEAs. Finally, the conclusions are drawn, and future work is discussed in Section \ref{sec:cl-fw}.

\section{Preliminary Studies and Related Works}
\label{sec:Preliminaries-and-Related}
In this section, we will briefly describe the definition of the LSMOP. Then we will discuss existing methods to solve LSMOPs and related works that motivate our methods.

\subsection{Large-scale Multiobjective Optimization}
Large-scale multiobjective optimization problems can be mathematically formulated as follows:
\begin{equation}
\begin{aligned}
\textbf{minimize}\ \boldsymbol F(\boldsymbol x) =& (f_1(\boldsymbol x), f_2(\boldsymbol x),\dots,f_m(\boldsymbol x))\\
&s.t.\ \boldsymbol x \in \Omega
\end{aligned}
\end{equation}
where $\boldsymbol x=(x_1,x_2,...,x_d)$ is $d$-dimensional decision vector, $\boldsymbol F=(f_1,f_2,\dots,f_m)$ is $m$-dimensional objective vector. It is worth noting that large-scale optimization problems consider the dimension in $d$ to be greater than $100$ \cite{RN85}.
\par
Suppose $\boldsymbol x_1$ and $\boldsymbol x_2$ are two solutions of an MOP, solution $\boldsymbol x_1$ is known to Pareto dominate solution $\boldsymbol x_2$ (denoted as $\boldsymbol x_1 \prec \boldsymbol x_2$), if and only if $f_i(\boldsymbol x_1) \leqslant f_i(\boldsymbol x_2) (\forall i = 1,\dots,m)$ and there exists at least one objective $f_j (j \in \{1, 2, \dots , m\})$ satisfying $f_j(\boldsymbol x_1) < f_j(\boldsymbol x_2)$. The collection of all the Pareto optimal solutions in the decision space is called the Pareto optimal set (POS), and the projection of POS in the objective space is called the Pareto optimal front (POF).
\subsection{Related Works}
Many algorithms have been developed for solving LSMOPs. These large-scale MOEAs could be categorized into three types: decision variable grouping based, decision space reduction based, and novel search strategy based approaches \cite{tian2021evolutionary}.
\par
\subsubsection{Decision Variable Grouping Based Large-scale MOEAs}
CCGDE3, proposed by Antonio \emph{et al.}, was the first MOEA based on decision variable grouping for solving LSMOPs. Ma \textit{et al.} \cite{RN106} proposed MOEA/DVA based on decision variable analysis, in which the decision vectors were decomposed into two groups via an interdependence variable analysis. Zhang \emph{et al.} \cite{RN82} proposed a large-scale evolutionary algorithm (LMEA) based on the decision variable clustering approach.
\subsubsection{Decision Space Reduction Based Large-scale MOEAs}
MOEA/PSL proposed by Tian \emph{et al.} in \cite{9047876} is a typical algorithm based on decision space reduction. The algorithm trained a restricted Boltzmann machine according to the binary vectors in an unsupervised manner. A weighted optimization framework (WOF) proposed by Zille \emph{et al.} \cite{RN89} introduced a problem transformation scheme that can be used to reduce the dimensionality of the search space.
\subsubsection{Novel Search Strategy Based Large-scale MOEAs}
In this category, many novel search strategies, such as improved crossover operators and probabilistic prediction model \cite{YI2020470, RN257} were suggested. LMOCSO proposed by Tian \emph{et al.} in \cite{RN85} used a competitive group optimizer to solve the LSMOP. In addition to GMOEA \cite{RN90} introduced before, Wang \emph{et al.} \cite{RN219} proposed a generative adversarial network-based manifold interpolation framework to generate high-quality solutions for improving the optimization performance.

\begin{figure*}[ht]
    \centering
    \includegraphics[width=0.9\hsize]{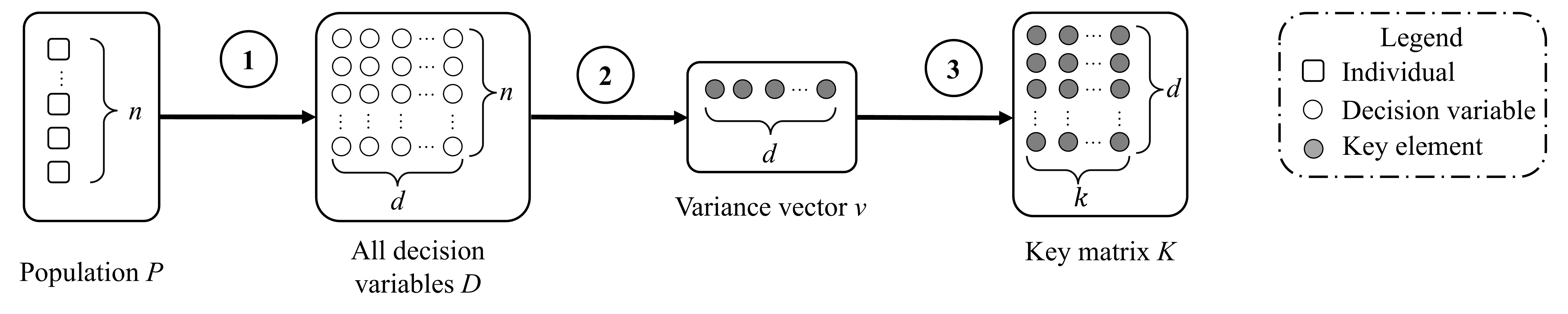}
    \caption{Process of calculating the Key matrix $K$. 1: $n$ $d$-dimensional individuals form the $n\times d$ decision variable matrix $D$; 2: Calculate the variance by column to get the $1\times d$ variance vector $\boldsymbol d$; 3: Construct the Key matrix $K$.}
    \label{fig:Key}
\end{figure*}

\subsection{Attention Mechanism}
Attention mechanism \cite{10.5555/3295222.3295349} was first used in the field of natural language processing (NLP) on machine translation tasks. The essence of the attention mechanism is to find the relationship between inputs spontaneously through the weight matrix. There are three separate matrices $Q$, $K$, and $V$ in the attention mechanism. Specifically, they result from different linear transformations of a set of input $X$. First, two representation matrices of inputs, $Q$ and $K$,  are needed to calculate attention between different inputs. Then the similarity of $Q$ and $K$ is calculated as the attention matrix. Finally, the algorithm will use the weight on another linear transformation of inputs $X$, which is the matrix $V$. The calculation for attention can be unified into a single function:
\begin{equation}
\begin{aligned}
\label{equ:attention}
\text{Attention}(Q, K, V) = \text{softmax} (\frac{Q \cdot K^{\top}}{\sqrt{d_k}})\cdot V
\end{aligned}
\end{equation}
The attention mechanism performs well due to its competitive performance and tremendous potential in processing text and images with large dimensions \cite{DBLP:journals/corr/abs-2012-12556}. Considering the large-scale decision variables of LSMOP, different decision variables have different effects on the optimization results, so different degrees of exploration and exploitation on different decision variables are required. Therefore, balancing exploration and exploitation from the perspective of decision variables is critical to solving the LSMOP. This paper proposes an algorithm for solving large-scale multiobjective optimization problems via the attention mechanism.
\section{Large-scale Multiobjective Optimization Problem via Attention Mechanism (LMOAM)}
\label{sec:LMOAM}
\begin{figure*}[ht]
    \centering
    \includegraphics[width=0.9\hsize]{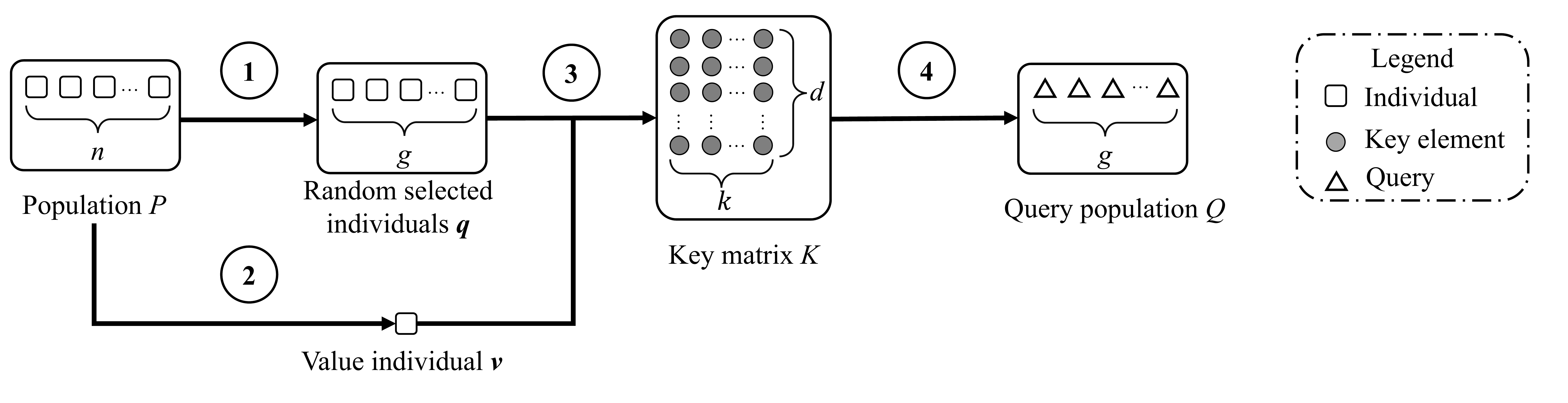}
    \caption{Process of calculating Query population $Q$. 1: Randomly select $g$ solutions; 2: Select value individual $\boldsymbol v$ with the largest crowding distance on the first front; 3: Multiply each decision vector $\boldsymbol q$ and $\boldsymbol v$ by the matrix $K$; 4: Construct Query $\boldsymbol q_a[k]=\boldsymbol v_a[k] / \boldsymbol q_a[k]$;}
    \label{fig:Query}
\end{figure*}

\subsection{Overview}
\par
Inspired by the attention function proposed in \cite{10.5555/3295222.3295349}, the attention mechanism is adopted for solving LSMOPs to balance exploration and exploitation from the perspective of decision variables. After the algorithm obtains the non-dominated solution set, the variance of the population in each dimension is calculated as the basis for decision variable extraction. Then the Key matrix $K$ is designed to extract representative decision variables according to their variance because the algorithm will assign approximate exploration and exploitation ratios to decision variables with approximate variances. And the Query population $Q$ is a set of weights designed to search for different degrees of exploration and exploitation on different decision variables extracted by the Key matrix. In each iteration, we select an individual with the largest crowding distance \cite{996017} on the first front of the POS as the Value $v$ to search the Query. The algorithm will balance exploration and exploitation from the perspective of decision variables on the Value individual $v$. The solution generated during the searching for Query and the evolved solution guided by Query will be added to the population pool as a newly generated population.
\par
In general, firstly, we calculate the variance of the whole population on each decision variable and construct the Key matrix $K$. Secondly, we use the Key matrix to construct a set of Query based on the population and search for a new set of Query via evolutionary algorithms. Thirdly, the algorithm multiplies the Query with the Key matrix to obtain a set of weights (attention). Finally, the algorithm assigns the set of attention to the individual $\boldsymbol v$ to obtain a new population.
\par
In our design, higher attention indicates exploration, lower attention indicates exploitation, and calculated attention on decision variables will guide the individual $v$ to perform exploration or exploitation on the decision variable level.
\subsection{Attention Mechanism}
\begin{figure}[ht]
    \centering
    \includegraphics[width=0.9\hsize]{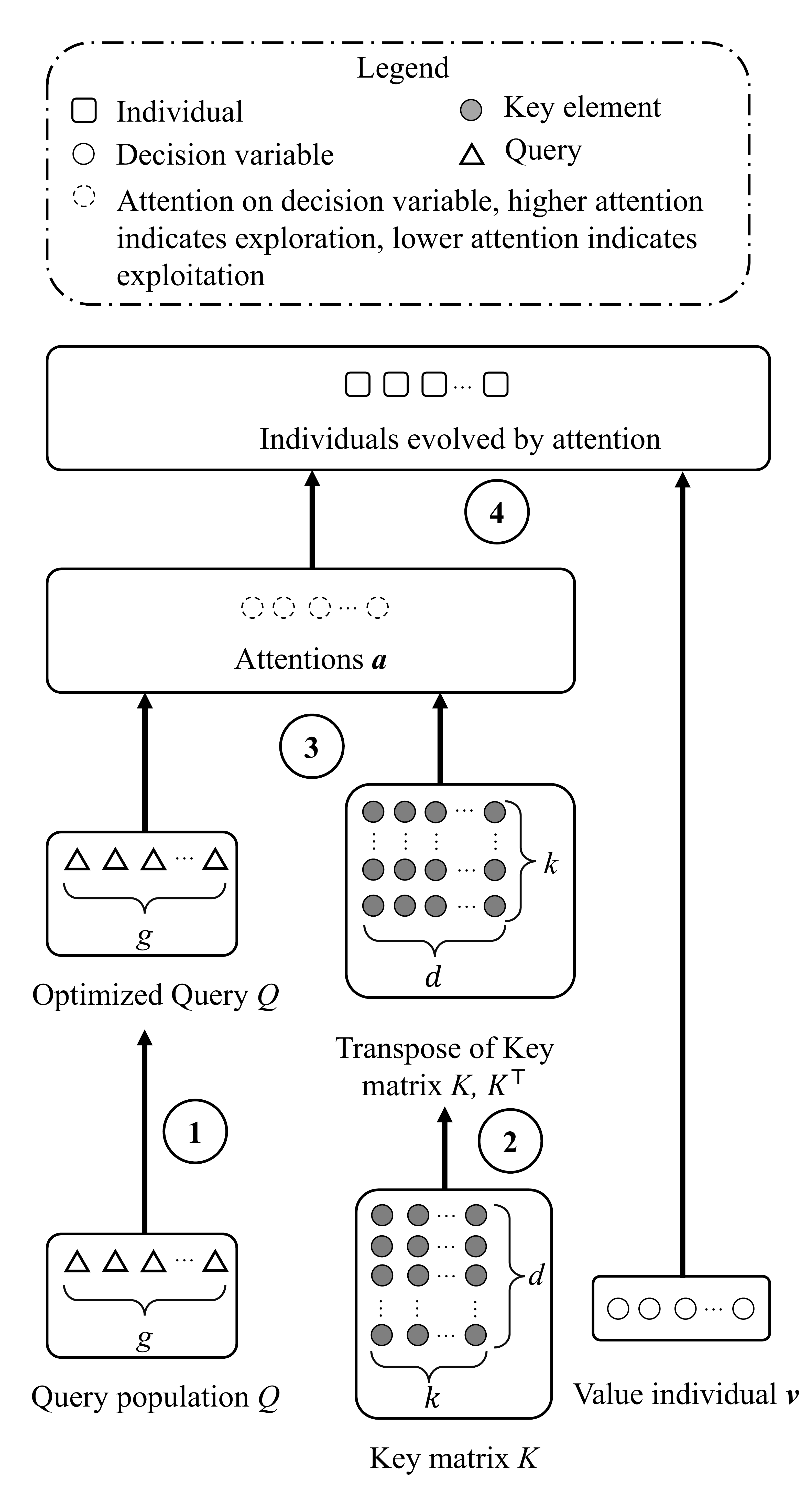}
    \caption{Basic process of Large-scale multiobjective optimization algorithm via attention mechanism. 1: Optimize Query population via evolutionary algorithm; 2: Transpose the Key matrix to obtain $K^{\top}$ to restore the full attention; 3: Multiply the Query by the matrix $K^{\top}$ to get the attention vector $\boldsymbol a$; 4: Obtain new individual $\boldsymbol v' = \boldsymbol a \circ \boldsymbol v$ and add it to the population.}
    \label{fig:LMOAM}
\end{figure}

\begin{algorithm}
    \caption{Query and Key}
    \label{alg:querykey}
    \KwIn{
        Population $P$; Dimension of Query $k$; Size of Queries $g$; Value individual $\boldsymbol v$.
    }
    \KwOut{A set of Queries $Q$; the Key matrix $K$.}
    Initializaiton \;
    $n \gets $ Size of population $P$\; 
    $d \gets $ Size of decision variable \;
    $D \gets $ Concatenation of population decision variables\;
    \For {$ i \gets 1$ to $d$} {
        $\boldsymbol d[i] \gets \text{Var}(D[:,i])$ \;
    }
    \For {$ i \gets 1$ to $d$} {
        $\boldsymbol d[i] \gets \frac{\boldsymbol d[i]-\min(\boldsymbol d)}{\max(\boldsymbol d) - \min(\boldsymbol d)}$ \;
    }
    $K \gets $ Empty matrix with the size of $d\times k$ \;
    \For {$K[i,j]$} {
        \eIf {$ j \times 1/k  \leq  \boldsymbol d[i] < (j+1) \times 1/k $}
        {
            $K[i,j] \gets 1$ \;
        } {
            $K[i,j] \gets 0$ \;
        }
    }
    \For {$ i \gets 1$ to $g$} {
        $\boldsymbol q \gets $ Randomly select an individual from $P$ \;
        $\boldsymbol q_a \gets \boldsymbol q \times K$, $\boldsymbol v_a \gets \boldsymbol v \times K$ \;
        \For {$ j \gets 1$ to $k$} {
            $\boldsymbol q_a[k] \gets \boldsymbol v_a[k] / \boldsymbol q_a[k]$ \;
        }
        $Q \gets Q \cup  \{ \boldsymbol q_a \}$ \;
    }
\end{algorithm}

In evolutionary optimization, if multiple individuals have a large variance in a certain decision variable, further exploration is required in this decision variable. If the variance is small, further exploitation is required for a finer search on this decision variable. The purpose of the Key matrix $K$ is to perform decision variable extraction. Therefore, we calculate the variance of the entire population in each decision variable as the calculation basis for extraction and group decision variables according to the variance.
\par
First, for a large-scale multiobjective optimization problem with $d$ decision variables, given a population $P$ with the size of $n$, the integrated decision variables are denoted as an $n \times d$ matrix $D$, each row of the matrix $D$ is an individual. The method of obtaining the Key matrix $K$ is to calculate the variance column by column on the $D$ matrix, that is, to calculate the variance of all populations on different decision variables, and obtain a $ 1 \times d$ decision variable variance vector $\boldsymbol d$. Then we normalize the variance vector $\boldsymbol d$ and construct the Key matrix $K$ based on the vector $\boldsymbol d$. $K$ is a $d \times k$ matrix, in which $k$ is the dimension of Query and is a hyperparameter.

The way of calculating variance vector $\boldsymbol d$ and constructing the Key matrix $K$ is given as following,

\begin{equation}
\begin{aligned}
\label{equ:varianceV}
\boldsymbol d:  \boldsymbol d[i] = \text{Var}(D[:,i])
\end{aligned}
\end{equation}

\noindent where $D[:,i]$ denotes the vertical cross section of $D$ with horizontal coordinate i, which is the $i$-th column of $D$. For the matrix of $K$, first, we make $k$ copies of the transpose of variance vector $\boldsymbol k$. Then for the $i$-th $(i=0,1,\dots,k-1)$ vector $\boldsymbol k[i]$, elements with values between $i \times 1/k$ and $(i+1) \times 1/k$ are recorded as 1, and the rest are recorded as 0. Finally $k$ vectors are normalized and concatenated into the $K$ matrix. The process of calculating the Key matrix is illustrated in in Fig. \ref{fig:Key}.

As for the Query. Firstly, we choose an individual $\boldsymbol v$ as the value individual from the set of solutions with the largest crowding distance on the first front. Then select other individuals $\boldsymbol q$ from the population. Afterwards, we multiply the decision vector $\boldsymbol q$ and $\boldsymbol v$ by the matrix $K$ to obtain the $\boldsymbol q_a$ and $\boldsymbol v_a$, then calculate the ratio of each dimension of $\boldsymbol q_a$ to each dimension of $\boldsymbol v_a$ to obtain a set of initialized Queries, that is, $\boldsymbol q_a[k]=\boldsymbol v_a[k] / \boldsymbol q_a[k]$. The illustration for this process is given in Fig. \ref{fig:Query}. The calculation steps for the Query and the Key matrix are given in Algorithm \ref{alg:querykey};
\subsection{Large-scale Multiobjective Optimization Algorithm via Attention Mechanism (LMOAM)}

\begin{algorithm}
    \caption{LMOAM}
    \label{alg:LMOAM}
    \KwIn{
        Dimension of Query $k$, Size of Queries $m$, Function Evaluation $e$.
    }
    \KwOut{Solution population $P$.}
    Initialization \;
    $P \gets $ Random initial population \;
    \While{Total function evaluations are not used} {
        $\boldsymbol v \gets $ Select a Value individual \;
        $Q, K \gets$ \textbf{Query and Key} ($P, k, m, \boldsymbol v$) \;
        \While{Function evaluations $e$ are not used}{
            $P' \gets \emptyset$ \;
            \For {$\boldsymbol q_m$ in $Q$} {
                $\boldsymbol a \gets \boldsymbol q_m \cdot K ^{\top}$ \;
                $\boldsymbol v' \gets \boldsymbol a \circ \boldsymbol v$ \;
                $P' \gets P' \cup \{ \boldsymbol v'\}$ \;
            }
            Evaluate $Q$ according to the evaluation of $P'$\;
            $Q \gets $ Evolutionary algorithm($Q$) \;
        }
        $P \gets P \cup P'$ \;
        $P \gets $ \textbf{Optimization Algorithm}($P, e$) \;
    }
\end{algorithm}

After obtaining a set of initialized Queries, we use an evolutionary algorithm to perform an iterative search. And the evaluation of Query is based on the individual's evaluation obtained after Query guides Value individual $v$ to explore or exploit on different decision variables. Specifically, we multiply the Query by the transpose of the Key matrix $K$ to get the attention vector $\boldsymbol a$, then take the Hadamard product of the attention $\boldsymbol a$ and $\boldsymbol v$ to get a new individual $\boldsymbol v'$ as follows:

\begin{equation}
\begin{aligned}
\label{equ:drtov}
\boldsymbol a & = \boldsymbol q \cdot K ^{\top}  \\
\boldsymbol v' & = \boldsymbol a \circ \boldsymbol v
\end{aligned}
\end{equation}
where  $\circ$ is the Hadamard product. Each Query can generate a new solution, and a group of Queries will produce a new population. Therefore, the evaluation value of Query is the objective function value of the solution generated by the corresponding Query. Afterward, a traditional evolutionary algorithm can be conducted on Queries to obtain a new set of Query. The main steps of the algorithm are given in Algorithm \ref{alg:LMOAM}, and the illustration for obtaining new individuals according to the attention is presented in Fig. \ref{fig:LMOAM}.

\section{Experimental Studies}
\label{sec:exp}

\begin{table*}[htbp]
\footnotesize
  \centering
  \caption{Statistics of IGD Values Achieved by eight Compared Algorithms on 36 Test Instances}
    \begin{tabular}{cccccccccc}
    \toprule
    Problem & D     & NSGAII & CCGDE3 & MOEADVA & LMEA  & MOEAPSL & LMOCSO & GMOEA & LMOAM (ours) \\
    \midrule
    \multirow{4}[2]{*}{LSMOP1} & \multicolumn{1}{r}{100} & 2.64e-01+ & 4.05e+00- & 1.78e+00- & \textbf{1.76e-01+} & 3.42e-01+ & 5.78e-01- & 2.35e-01+ & 4.78E-01 \\
          & \multicolumn{1}{r}{500} & 1.80e+00- & 7.71e+00- & \textbf{4.22e-01+} & 1.04e+01- & 8.61e-01= & 1.34e+00- & 1.91e+00- & 8.46E-01 \\
          & \multicolumn{1}{r}{1000} & 4.06e+00- & 7.20e+00- & \textbf{1.42e-01+} & 1.14e+01- & 2.59e+00- & 1.31e+00- & 7.47e+01- & 8.49E-01 \\
          & \multicolumn{1}{r}{5000} & 9.61e+00- & 1.02e+01- & 8.55e+01- & 8.55e+01- & 1.86e+00- & 1.47e+00- & 8.28e+01- & \textbf{8.46e-01} \\
    \midrule
    \multirow{4}[2]{*}{LSMOP2} & \multicolumn{1}{r}{100} & 1.73e-01= & 2.15e-01= & 1.75e-01= & \textbf{8.61e-02+} & 2.05e-01= & 1.30e-01+ & 1.62e-01= & 1.87E-01 \\
          & \multicolumn{1}{r}{500} & 6.30e-02= & 6.89e-02= & 5.63e-02= & 6.28e-02= & 6.17e-02= & \textbf{4.43e-02=} & 2.42e+00- & 5.85E-02 \\
          & \multicolumn{1}{r}{1000} & 4.62e-02= & 4.54e-02= & 4.10e-02= & 4.48e-02= & 4.53e-02= & \textbf{3.20e-02=} & 7.47e+01- & 4.18E-02 \\
          & \multicolumn{1}{r}{5000} & 3.75e-02= & 3.55e-02= & 8.58e+01- & 8.53e+01- & 3.42e-02= & \textbf{2.35e-02=} & 8.29e+01- & 3.22E-02 \\
    \midrule
    \multirow{4}[2]{*}{LSMOP3} & \multicolumn{1}{r}{100} & 1.11e+00- & 1.12e+01- & 2.27e+01- & 2.31e+00- & \textbf{7.75e-01+} & 9.35e+00- & 1.06e+00- & 8.61E-01 \\
          & \multicolumn{1}{r}{500} & 8.11e+00- & 1.59e+01- & 4.67e+00- & 1.03e+03- & 1.08e+00- & 1.17e+01- & 2.28e+00- & \textbf{8.61e-01} \\
          & \multicolumn{1}{r}{1000} & 1.41e+01- & 2.04e+01- & 2.28e+00- & 1.72e+02- & 8.61e-01= & 1.20e+01- & 7.48e+01- & \textbf{8.61e-01} \\
          & \multicolumn{1}{r}{5000} & 2.16e+01- & 2.04e+01- & 8.57e+01- & 8.56e+01- & 8.61e-01= & 1.34e+01- & 8.33e+01- & \textbf{8.61e-01} \\
    \midrule
    \multirow{4}[2]{*}{LSMOP4} & \multicolumn{1}{r}{100} & 4.35e-01= & 5.21e-01- & 4.47e-01- & \textbf{2.41e-01+} & 3.13e-01+ & 3.76e-01= & 3.01e-01+ & 3.94E-01 \\
          & \multicolumn{1}{r}{500} & 1.82e-01= & 2.06e-01= & \textbf{9.11e-02+} & 2.02e-01= & 1.74e-01= & 1.47e-01= & 2.73e+00- & 1.96E-01 \\
          & \multicolumn{1}{r}{1000} & 1.13e-01= & 1.26e-01= & \textbf{4.68e-02+} & 1.18e-01= & 1.00e-01= & 8.74e-02= & 7.50e+01- & 1.17E-01 \\
          & \multicolumn{1}{r}{5000} & 4.27e-02= & 4.18e-02= & 8.54e+01- & 8.54e+01- & 4.55e-02= & \textbf{3.15e-02=} & 8.28e+01- & 4.11E-02 \\
    \midrule
    \multirow{4}[2]{*}{LSMOP5} & \multicolumn{1}{r}{100} & 3.22e-01= & 3.45e+00- & 2.05e+00- & 8.36e+00- & \textbf{2.72e-01=} & 6.65e-01- & 2.80e-01= & 3.16E-01 \\
          & \multicolumn{1}{r}{500} & 3.79e+00- & 1.41e+01- & 9.09e-01= & 1.75e+01- & \textbf{5.41e-01+} & 2.89e+00- & 2.41e+00- & 9.46E-01 \\
          & \multicolumn{1}{r}{1000} & 7.69e+00- & 1.49e+01- & \textbf{4.05e-01+} & 2.03e+01- & 5.41e-01+ & 3.13e+00- & 7.54e+01- & 9.46E-01 \\
          & \multicolumn{1}{r}{5000} & 2.08e+01- & 1.83e+01- & 8.55e+01- & 8.61e+01- & \textbf{9.46e-01=} & 3.08e+00- & 8.34e+01- & \textbf{9.46e-01} \\
    \midrule
    \multirow{4}[2]{*}{LSMOP6} & \multicolumn{1}{r}{100} & 1.21e+00+ & 1.32e+03- & 7.75e+02- & 2.09e+01- & \textbf{1.12e+00+} & 3.15e+00- & 2.82e+00- & 1.31E+00 \\
          & \multicolumn{1}{r}{500} & 4.05e+01- & 9.75e+03- & 1.54e+02- & 3.17e+04- & \textbf{1.31e+00+} & 1.72e+02- & 2.56e+00- & 1.64E+00 \\
          & \multicolumn{1}{r}{1000} & 1.37e+03- & 1.30e+04- & 3.60e+01- & 2.45e+02- & 1.31e+00= & 3.45e+02- & 7.54e+01- & \textbf{1.31e+00} \\
          & \multicolumn{1}{r}{5000} & 1.54e+04- & 2.24e+04- & 8.51e+01- & 8.60e+01- & 1.79e+02- & 2.43e+02- & 8.27e+01- & \textbf{1.70e+00} \\
    \midrule
    \multirow{4}[2]{*}{LSMOP7} & \multicolumn{1}{r}{100} & 1.87e+00- & 2.70e+00- & 5.07e+01- & 2.82e+00- & 1.24e+00- & 1.38e+00- & 1.14e+00- & \textbf{8.28e-01} \\
          & \multicolumn{1}{r}{500} & 1.26e+00- & 1.29e+00- & 2.51e+00- & 5.35e+00- & 1.20e+00- & 1.11e+00- & 2.22e+00- & \textbf{8.97e-01} \\
          & \multicolumn{1}{r}{1000} & 1.11e+00- & 1.10e+00- & 3.48e+00- & 3.56e+00- & 1.08e+00- & 1.02e+00- & 7.46e+01- & \textbf{9.03e-01} \\
          & \multicolumn{1}{r}{5000} & 9.73e-01- & 9.72e-01- & 8.61e+01- & 8.52e+01- & \textbf{8.50e-01+} & 9.59e-01= & 8.31e+01- & 9.19E-01 \\
    \midrule
    \multirow{4}[2]{*}{LSMOP8} & \multicolumn{1}{r}{100} & 3.61e-01- & 8.07e-01- & 6.85e-01- & 2.72e-01= & 3.63e-01- & 6.68e-01- & 3.47e-01- & \textbf{2.45e-01} \\
          & \multicolumn{1}{r}{500} & 7.66e-01- & 9.38e-01- & 4.50e-01- & 6.45e-01- & 5.78e-01- & 5.42e-01- & 2.19e+00- & \textbf{3.20e-01} \\
          & \multicolumn{1}{r}{1000} & 7.84e-01- & 9.58e-01- & \textbf{2.15e-01+} & 7.24e-01- & 5.49e-01- & 5.36e-01- & 7.50e+01- & 3.31E-01 \\
          & \multicolumn{1}{r}{5000} & 9.51e-01- & 9.51e-01- & 8.53e+01- & 8.58e+01- & 5.71e-01- & 5.21e-01- & 8.30e+01- & \textbf{3.25e-01} \\
    \midrule
    \multirow{4}[2]{*}{LSMOP9} & \multicolumn{1}{r}{100} & 1.19e+00= & 2.05e+01- & 1.96e+01- & 1.03e+00+ & 1.52e+00- & \textbf{5.58e-01+} & 1.16e+00= & 1.15E+00 \\
          & \multicolumn{1}{r}{500} & 3.14e+00- & 6.63e+01- & 2.81e+00- & 1.38e+02- & 1.54e+00- & 1.70e+00- & 2.59e+00- & \textbf{1.15e+00} \\
          & \multicolumn{1}{r}{1000} & 6.92e+00- & 7.66e+01- & \textbf{6.89e-01+} & 2.76e+02- & 1.51e+01- & 5.48e+01- & 7.51e+01- & 1.15E+00 \\
          & \multicolumn{1}{r}{5000} & 2.84e+01- & 8.44e+01- & 8.52e+01- & 8.54e+01- & \textbf{1.17e+00=} & 7.92e+01- & 8.29e+01- & 1.22E+00 \\
    \midrule
    (+/-/=) &       & {2/24/10} & {0/29/7} & {7/25/4} & {4/27/5} & {8/14/14} & {2/26/8} & {2/31/3} & \\
    \bottomrule
    \end{tabular}%
  \label{tab:IGD}%
\end{table*}%

\subsection{Algorithms in Comparison and Test Problems}
To empirically validate the performance of our proposed method, one representative MOEA NSGAII\cite{RN110} and six state-of-the-art large-scale MOEAs, CCGDE3 \cite{6557903}, MOEADVA\cite{7155533}, LMEA \cite{RN82}, MOEA/PSL \cite{9047876}, LMOCSO \cite{RN85} and GMOEA \cite{9082904} are compared on the LSMOP suite \cite{RN96}.  All the compared algorithms are implemented on PlatEMO \cite{RN92}.
\par
Each compared method is run 20 times on each test problem independently, and the Wilcoxon rank-sum test \cite{Haynes2013} is adopted to compare the results at a significance level of 0.05. Symbols "+" "-" and "=" in the tables indicate the compared algorithm is significantly better than, significantly worse than, and statistically tied by LMOAM.
\subsection{Parameter Settings and Performance Indicators}
\begin{enumerate}
\item Population Size: The population size $n$ is set to 300 for all test instances;
\item Termination Condition: The number of maximum evaluations $e$ is set to 100,000;
\item The function evaluation $e$ for optimization algorithm in Algorithm \ref{alg:LMOAM} is set to $0.05e$;
\item Dimension of Query $k$ is set to 5;
\item Size of Queries $g$ is set to 20;
\item According to \cite{RN92}, the parameters of compared algorithms are set according to their original publications.
\end{enumerate}

In the experiments, two widely used performance indicators, the inverted generational distance (IGD) \cite{RN98} and hypervolume (HV) \cite{1583625} are used for comparing the performance.

\subsection{Performance on LSMOP Problems}

\begin{table*}[htbp]
  \centering
  \caption{Statistics of HV Values Achieved by eight Compared Algorithms on 36 Test Instances}
    \begin{tabular}{cccccccccc}
    \toprule
    Problem & D     & NSGAII & CCGDE3 & MOEADVA & LMEA  & MOEAPSL & LMOCSO & GMOEA & LMOAM (ours) \\
    \midrule
    \multirow{4}[2]{*}{LSMOP1} & \multicolumn{1}{r}{100} & 5.45e-01+ & 0.00e+00- & 0.00e+00- & \textbf{6.27e-01+} & 4.00e-01+ & 1.91e-01= & 5.83e-01+ & 2.29E-01 \\
          & \multicolumn{1}{r}{500} & 0.00e+00- & 0.00e+00- & \textbf{3.31e-01+} & 0.00e+00- & 9.09e-02= & 0.00e+00- & 1.30e-02- & 1.01E-01 \\
          & \multicolumn{1}{r}{1000} & 0.00e+00- & 0.00e+00- & \textbf{7.03e-01+} & 0.00e+00- & 0.00e+00- & 0.00e+00- & 0.00e+00- & 9.85E-02 \\
          & \multicolumn{1}{r}{5000} & 0.00e+00- & 0.00e+00- & 0.00e+00- & 0.00e+00- & 0.00e+00- & 0.00e+00- & 0.00e+00- & \textbf{9.94e-02} \\
    \midrule
    \multirow{4}[2]{*}{LSMOP2} & \multicolumn{1}{r}{100} & 6.79e-01= & 6.26e-01= & 6.66e-01= & \textbf{7.74e-01+} & 6.34e-01= & 7.34e-01+ & 7.04e-01= & 6.65E-01 \\
          & \multicolumn{1}{r}{500} & 8.05e-01= & 7.99e-01= & 8.05e-01= & 7.92e-01= & 8.04e-01= & \textbf{8.27e-01=} & 1.30e-02- & 8.12E-01 \\
          & \multicolumn{1}{r}{1000} & 8.25e-01= & 8.24e-01= & 8.24e-01= & 8.19e-01= & 8.24e-01= & \textbf{8.41e-01=} & 0.00e+00- & 8.30E-01 \\
          & \multicolumn{1}{r}{5000} & 8.40e-01= & 8.41e-01= & 0.00e+00- & 0.00e+00- & 8.40e-01= & \textbf{8.52e-01=} & 0.00e+00- & 8.45E-01 \\
    \midrule
    \multirow{4}[2]{*}{LSMOP3} & \multicolumn{1}{r}{100} & 0.00e+00- & 0.00e+00- & 0.00e+00- & 0.00e+00- & 9.09e-02= & 0.00e+00- & 0.00e+00- & \textbf{9.09e-02} \\
          & \multicolumn{1}{r}{500} & 0.00e+00- & 0.00e+00- & 0.00e+00- & 0.00e+00- & 0.00e+00- & 0.00e+00- & 1.30e-02- & \textbf{9.09e-02} \\
          & \multicolumn{1}{r}{1000} & 0.00e+00- & 0.00e+00- & 0.00e+00- & 0.00e+00- & 9.09e-02= & 0.00e+00- & 0.00e+00- & \textbf{9.09e-02} \\
          & \multicolumn{1}{r}{5000} & 0.00e+00- & 0.00e+00- & 0.00e+00- & 0.00e+00- & 9.09e-02= & 0.00e+00- & 0.00e+00- & \textbf{9.09e-02} \\
    \midrule
    \multirow{4}[2]{*}{LSMOP4} & \multicolumn{1}{r}{100} & 3.32e-01- & 2.49e-01- & 3.16e-01- & \textbf{5.59e-01+} & 4.71e-01+ & 4.34e-01= & 4.98e-01+ & 3.96E-01 \\
          & \multicolumn{1}{r}{500} & 6.58e-01= & 6.38e-01= & \textbf{7.68e-01+} & 6.21e-01= & 6.80e-01= & 7.11e-01= & 1.30e-02- & 6.65E-01 \\
          & \multicolumn{1}{r}{1000} & 7.44e-01= & 7.36e-01= & \textbf{8.15e-01+} & 7.31e-01= & 7.63e-01= & 7.80e-01= & 0.00e+00- & 7.46E-01 \\
          & \multicolumn{1}{r}{5000} & 8.27e-01= & 8.29e-01= & 0.00e+00- & 0.00e+00- & 8.26e-01= & \textbf{8.42e-01=} & 0.00e+00- & 8.31E-01 \\
    \midrule
    \multirow{4}[2]{*}{LSMOP5} & \multicolumn{1}{r}{100} & 2.17e-01- & 0.00e+00- & 0.00e+00- & 0.00e+00- & \textbf{3.98e-01+} & 1.98e-03- & 2.47e-01- & 3.46E-01 \\
          & \multicolumn{1}{r}{500} & 0.00e+00- & 0.00e+00- & 0.00e+00- & 0.00e+00- & \textbf{3.50e-01+} & 0.00e+00- & 1.30e-02- & 9.09E-02 \\
          & \multicolumn{1}{r}{1000} & 0.00e+00- & 0.00e+00- & 1.05e-01= & 0.00e+00- & \textbf{3.49e-01+} & 0.00e+00- & 0.00e+00- & 9.09E-02 \\
          & \multicolumn{1}{r}{5000} & 0.00e+00- & 0.00e+00- & 0.00e+00- & 0.00e+00- & \textbf{9.09e-02=} & 0.00e+00- & 0.00e+00- & \textbf{9.09e-02} \\
    \midrule
    \multirow{4}[2]{*}{LSMOP6} & \multicolumn{1}{r}{100} & 0.00e+00= & 0.00e+00= & 0.00e+00= & 0.00e+00= & 0.00e+00= & 0.00e+00= & 0.00e+00= & 0.00E+00 \\
          & \multicolumn{1}{r}{500} & 0.00e+00= & 0.00e+00= & 0.00e+00= & 0.00e+00= & 0.00e+00= & 0.00e+00= & \textbf{1.30e-02=} & 0.00E+00 \\
          & \multicolumn{1}{r}{1000} & 0.00e+00= & 0.00e+00= & 0.00e+00= & 0.00e+00= & 0.00e+00= & 0.00e+00= & 0.00e+00= & 0.00E+00 \\
          & \multicolumn{1}{r}{5000} & 0.00e+00= & 0.00e+00= & 0.00e+00= & 0.00e+00= & 0.00e+00= & 0.00e+00= & 0.00e+00= & 0.00E+00 \\
    \midrule
    \multirow{4}[2]{*}{LSMOP7} & \multicolumn{1}{r}{100} & 0.00e+00- & 0.00e+00- & 0.00e+00- & 0.00e+00- & 0.00e+00- & 0.00e+00- & 0.00e+00- & \textbf{8.90e-02} \\
          & \multicolumn{1}{r}{500} & 0.00e+00- & 0.00e+00- & 0.00e+00- & 0.00e+00- & 0.00e+00- & 0.00e+00- & 1.30e-02- & \textbf{8.14e-02} \\
          & \multicolumn{1}{r}{1000} & 0.00e+00- & 0.00e+00- & 0.00e+00- & 0.00e+00- & 0.00e+00- & 0.00e+00- & 0.00e+00- & \textbf{7.35e-02} \\
          & \multicolumn{1}{r}{5000} & 4.28e-02= & 4.37e-02= & 0.00e+00- & 0.00e+00- & 3.62e-02= & 4.90e-02= & 0.00e+00- & \textbf{7.58e-02} \\
    \midrule
    \multirow{4}[2]{*}{LSMOP8} & \multicolumn{1}{r}{100} & 2.79e-01- & 0.00e+00- & 0.00e+00- & 2.60e-01- & \textbf{3.74e-01=} & 6.39e-03- & 3.14e-01= & 3.36E-01 \\
          & \multicolumn{1}{r}{500} & 5.25e-02- & 5.13e-02- & 1.38e-01- & 5.19e-02- & 5.48e-02- & 7.24e-02- & 1.30e-02- & \textbf{3.55e-01} \\
          & \multicolumn{1}{r}{1000} & 6.77e-02- & 6.88e-02- & 3.03e-01= & 0.00e+00- & 8.13e-02- & 8.56e-02- & 0.00e+00- & \textbf{3.47e-01} \\
          & \multicolumn{1}{r}{5000} & 8.12e-02- & 8.13e-02- & 0.00e+00- & 0.00e+00- & 1.12e-01- & 9.90e-02- & 0.00e+00- & \textbf{3.54e-01} \\
    \midrule
    \multirow{4}[2]{*}{LSMOP9} & \multicolumn{1}{r}{100} & 1.38e-01= & 0.00e+00- & 0.00e+00- & 4.53e-02- & 9.38e-02- & 3.41e-02- & 1.44e-01= & \textbf{1.47e-01} \\
          & \multicolumn{1}{r}{500} & 0.00e+00- & 0.00e+00- & 0.00e+00- & 0.00e+00- & 9.11e-02- & 1.56e-02- & 1.30e-02- & \textbf{1.47e-01} \\
          & \multicolumn{1}{r}{1000} & 0.00e+00- & 0.00e+00- & 4.79e-02- & 0.00e+00- & 0.00e+00- & 0.00e+00- & 0.00e+00- & \textbf{1.47e-01} \\
          & \multicolumn{1}{r}{5000} & 0.00e+00- & 0.00e+00- & 0.00e+00- & 0.00e+00- & 1.17e-01= & 0.00e+00- & 0.00e+00- & \textbf{1.36e-01} \\
    \midrule
    (+/-/=) &       & {1/22/13} & {0/24/12} & {4/23/9} & {3/25/8} & {5/12/19} & {1/22/13} & {2/27/7} & \\
    \bottomrule
    \end{tabular}%
  \label{tab:HV}%
\end{table*}%

In this section, we compare our proposed LMOAM with other state-of-the-art large-scale MOEAs in Table \ref{tab:IGD} and \ref{tab:HV}. Generally, the proposed method LMOAM has better performance in convergence and diversity.
\par
As can be observed, LMOAM has achieved 14 out of 36 best results in Table \ref{tab:IGD} with respect to the IGD indicator. MOEA/DVA and MOEA/PSL have achieved seven and eight best results, respectively. LMOCSO and LMEA have achieved five and three best results, respectively. As for the indicator of HV in Table \ref{tab:HV}, our proposed method LMOAM achieved 17 best results out of 36 test instances. MOEA/DVA and MOEA/PSL have achieved four and five best results. Our proposed method mainly achieved the best results on LSMOP3, LSMOP7, LSMOP8, and LSMOP9.

\subsection{Convergence and Computation Time Analysis}
\begin{figure*}[htbp]
  \centering
  \subfloat[]{\includegraphics[width=0.29\hsize]{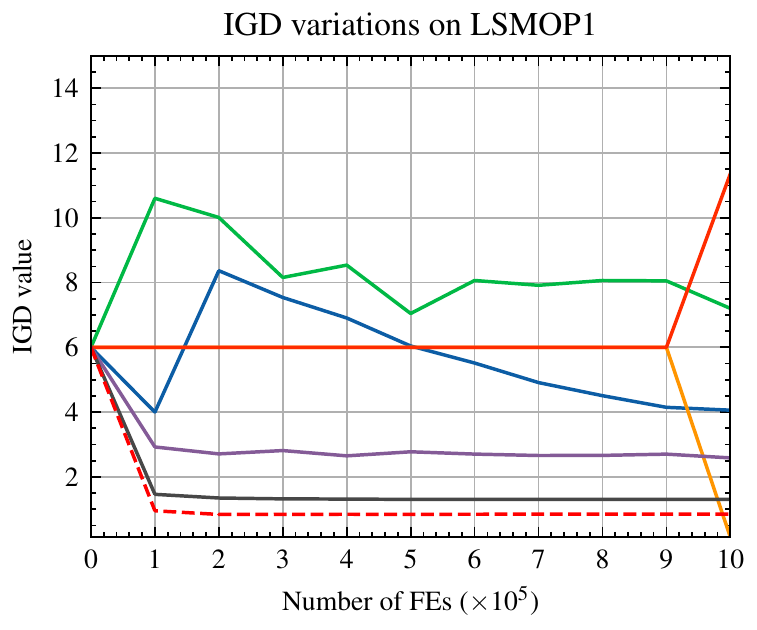}}
  \subfloat[]{\includegraphics[width=0.295\hsize]{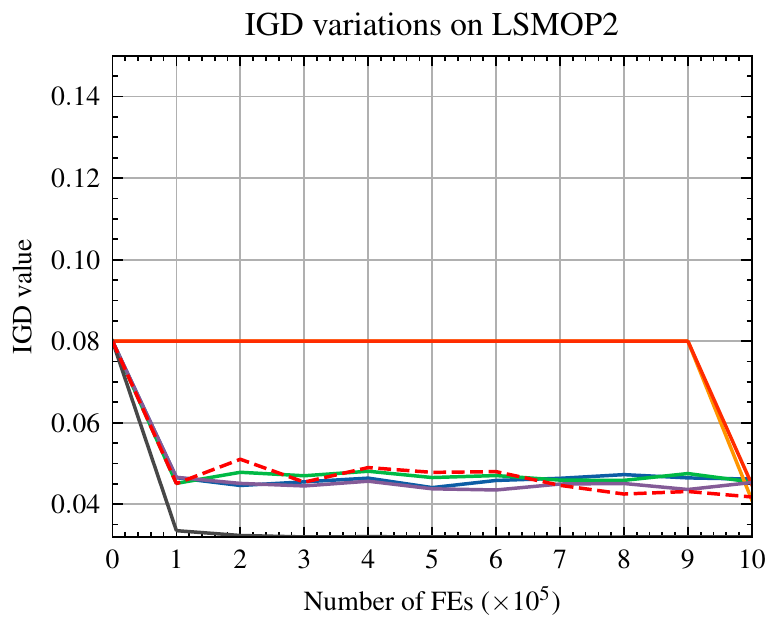}}
  \subfloat[]{\includegraphics[width=0.40\hsize]{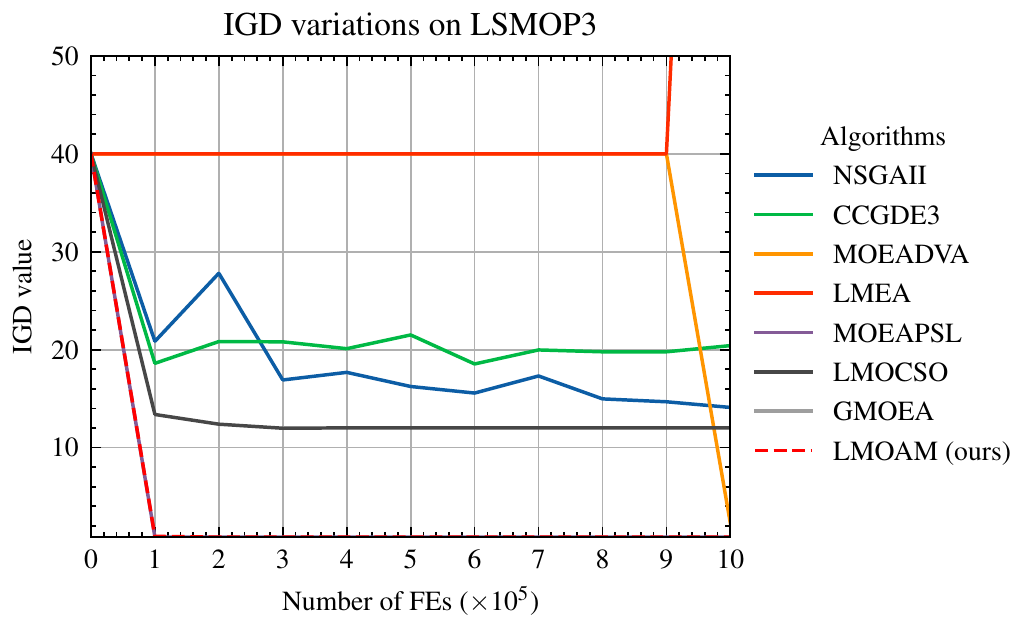}} \\
    \caption{Variations of IGD values achieved by compared algorithms on tri-objective LSMOP1 - LSMOP3 with 1,000 decision variables, respectively}
    \label{fig:result_rate}
\end{figure*}

\begin{figure*}[htbp]
  \centering
  \subfloat[]{\includegraphics[width=0.3\hsize]{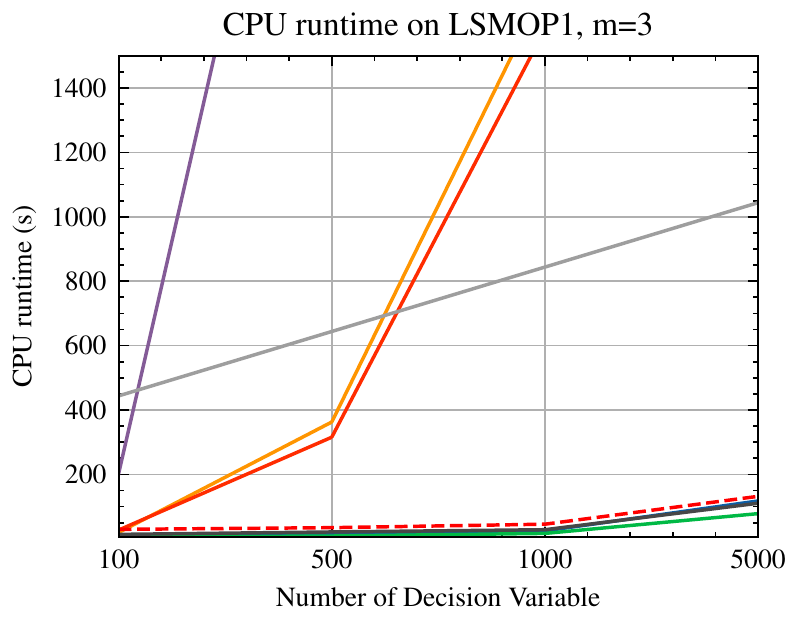}}
  \subfloat[]{\includegraphics[width=0.3\hsize]{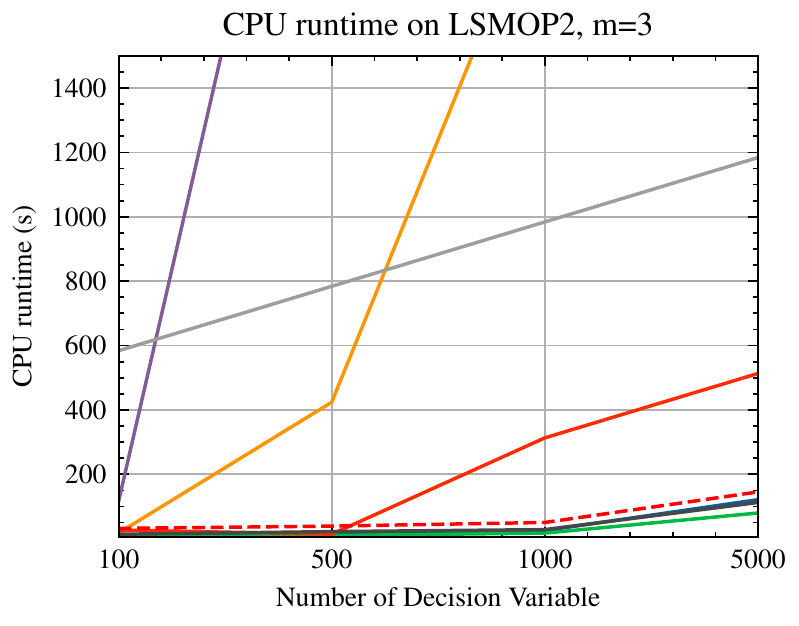}}
  \subfloat[]{\includegraphics[width=0.4\hsize]{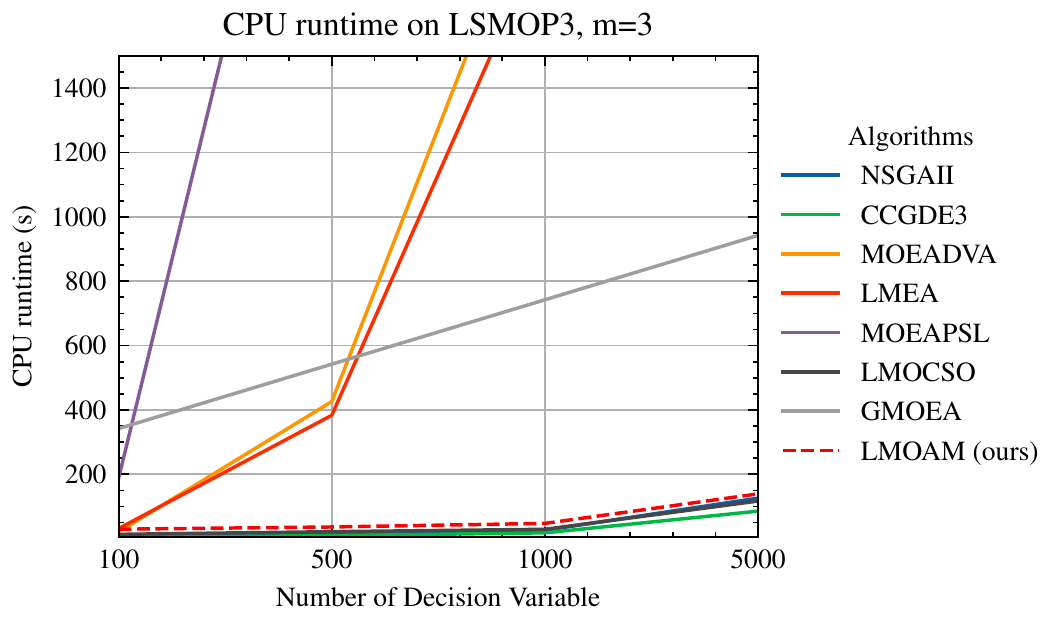}} \\
    \caption{The CPU running time of the compared algorithm in solving tri-objective LSMOPs, where FE=100,000, and the dimension of the decision variable varies from 100 to 5000.}
    \label{fig:time}
\end{figure*}

The convergence profiles of all compared algorithms on tri-objective LSMOP1 to LSMOP3 with 1,000 decision variables are reported in Fig. \ref{fig:result_rate}, and the average computation time on LSMOP1 to LMSOP3 is displayed in Fig. \ref{fig:time}. It can be observed from these figures that our proposed LMOAM has the fastest convergence rate on those two test instances, while its computation time is similar to that of CCGDE3 and NSGAII. Our algorithm has a fast convergence rate, and the computational cost of the algorithm does not increase exponentially with the increase of dimensions. 
\subsection{Discussion}
Based on the statistical results obtained from the above experiments, it can be observed that the proposed algorithm LMOAM has good performance in solving large-scale multiobjective optimization problems. The results of the IGD and HV metrics show that by balancing exploration and exploitation on different decision variables, the proposed algorithm outperforms several state-of-the-art algorithms in convergence and diversity. While ensuring the performance of the algorithm, our algorithm also has advantages in convergence and computational time. The experimental results verify that assigning attention to decision variables and optimizing accordingly is effective for solving LSMOPs.

\section{Conclusion and Future works}
\label{sec:cl-fw}
In this paper, we propose an algorithm that calculates attention to decision variables for solving LSMOPs. The key difficulty of the LSMOP is the curse of dimensionality caused by large-scale decision variables. And balancing exploration and exploitation in large-scale decision variables is an important method to solve the curse of dimensionality. Inspired by the attention mechanism, the algorithm assigns attention to different decision variables and balances the exploration and exploitation from the perspective of the decision variable. The experimental results validate that our algorithm has an advantage in performance compared with state-of-the-art large-scale MOEAs. In our future work, other machine learning techniques \cite{jiang2020fast} \cite{jiang2020individual} will be considered to solve LSMOPs.

\section*{Acknowledgment}
This work was supported in part by the High-end Foreign Expert Introduction Program of the Ministry of science and technology of China, the National Natural Science Foundation of China under Grant 61673328, and in part by the Collaborative Project Foundation of Fuzhou-Xiamen-Quanzhou Innovation Demonstration Zone under Grant 3502ZCQXT202001.


\bibliographystyle{IEEEtran}
\bibliography{LMOAM_bibtex}

\begin{thebibliography}{10}
\providecommand{\url}[1]{#1}
\csname url@samestyle\endcsname
\providecommand{\newblock}{\relax}
\providecommand{\bibinfo}[2]{#2}
\providecommand{\BIBentrySTDinterwordspacing}{\spaceskip=0pt\relax}
\providecommand{\BIBentryALTinterwordstretchfactor}{4}
\providecommand{\BIBentryALTinterwordspacing}{\spaceskip=\fontdimen2\font plus
\BIBentryALTinterwordstretchfactor\fontdimen3\font minus
  \fontdimen4\font\relax}
\providecommand{\BIBforeignlanguage}[2]{{%
\expandafter\ifx\csname l@#1\endcsname\relax
\typeout{** WARNING: IEEEtran.bst: No hyphenation pattern has been}%
\typeout{** loaded for the language `#1'. Using the pattern for}%
\typeout{** the default language instead.}%
\else
\language=\csname l@#1\endcsname
\fi
#2}}
\providecommand{\BIBdecl}{\relax}
\BIBdecl

\bibitem{tian2021evolutionary}
Y.~Tian, L.~Si, X.~Zhang, R.~Cheng, C.~He, K.~c. Tan, and Y.~Jin,
  ``Evolutionary large-scale multi-objective optimization: A survey,''
  \emph{ACM Computing Surveys}, vol.~1, no.~1, 2021.

\bibitem{8618599}
C.~Pizzuti and A.~Socievole, ``Multiobjective optimization and local merge for
  clustering attributed graphs,'' \emph{IEEE Transactions on Cybernetics},
  vol.~50, no.~12, pp. 4997--5009, 2020.

\bibitem{8482477}
Y.~Tian, S.~Yang, L.~Zhang, F.~Duan, and X.~Zhang, ``A surrogate-assisted
  multiobjective evolutionary algorithm for large-scale task-oriented pattern
  mining,'' \emph{IEEE Transactions on Emerging Topics in Computational
  Intelligence}, vol.~3, no.~2, pp. 106--116, 2019.

\bibitem{RN258}
\BIBentryALTinterwordspacing
S.~Nguyen, M.~Zhang, M.~Johnston, and K.~C. Tan, ``Learning iterative
  dispatching rules for job shop scheduling with genetic programming,''
  \emph{The International Journal of Advanced Manufacturing Technology},
  vol.~67, no.~1, pp. 85--100, 2013. [Online]. Available:
  \url{https://doi.org/10.1007/s00170-013-4756-9}
\BIBentrySTDinterwordspacing

\bibitem{TAN2005485}
\BIBentryALTinterwordspacing
W.~Tan, F.~Lu, A.~Loh, and K.~Tan, ``Modeling and control of a pilot ph plant
  using genetic algorithm,'' \emph{Engineering Applications of Artificial
  Intelligence}, vol.~18, no.~4, pp. 485--494, 2005. [Online]. Available:
  \url{https://www.sciencedirect.com/science/article/pii/S0952197604001691}
\BIBentrySTDinterwordspacing

\bibitem{hong2021evolutionary}
W.-J. Hong, P.~Yang, and K.~Tang, ``Evolutionary computation for large-scale
  multi-objective optimization: A decade of progresses,'' \emph{International
  Journal of Automation and Computing}, pp. 1--15, 2021.

\bibitem{996017}
K.~Deb, A.~Pratap, S.~Agarwal, and T.~Meyarivan, ``A fast and elitist
  multiobjective genetic algorithm: Nsga-ii,'' \emph{IEEE Transactions on
  Evolutionary Computation}, vol.~6, no.~2, pp. 182--197, 2002.

\bibitem{RN135}
M.~Jiang, Z.~Huang, L.~Qiu, W.~Huang, and G.~G. Yen, ``Transfer learning-based
  dynamic multiobjective optimization algorithms,'' \emph{IEEE Transactions on
  Evolutionary Computation}, vol.~22, no.~4, pp. 501--514, 2018.

\bibitem{6848830}
V.~A. Shim, K.~C. Tan, and H.~Tang, ``Adaptive memetic computing for
  evolutionary multiobjective optimization,'' \emph{IEEE Transactions on
  Cybernetics}, vol.~45, no.~4, pp. 610--621, 2015.

\bibitem{RN106}
X.~Ma, F.~Liu, Y.~Qi, X.~Wang, L.~Li, L.~Jiao, M.~Yin, and M.~Gong, ``A
  multiobjective evolutionary algorithm based on decision variable analyses for
  multiobjective optimization problems with large-scale variables,'' \emph{IEEE
  Transactions on Evolutionary Computation}, vol.~20, no.~2, pp. 275--298,
  2016.

\bibitem{RN89}
H.~Zille, H.~Ishibuchi, S.~Mostaghim, and Y.~Nojima, ``A framework for
  large-scale multiobjective optimization based on problem transformation,''
  \emph{IEEE Transactions on Evolutionary Computation}, vol.~22, no.~2, pp.
  260--275, 2018.

\bibitem{RN222}
H.~Hong, K.~Ye, M.~Jiang, and K.~C. Tan, ``Solving large-scale multi-objective
  optimization via probabilistic prediction model,'' in \emph{Evolutionary
  Multi-Criterion Optimization}, H.~Ishibuchi, Q.~Zhang, R.~Cheng, K.~Li,
  H.~Li, H.~Wang, and A.~Zhou, Eds.\hskip 1em plus 0.5em minus 0.4em\relax
  Springer International Publishing, 2021, Conference Proceedings, pp.
  605--616.

\bibitem{6557903}
L.~M. Antonio and C.~A.~C. Coello, ``Use of cooperative coevolution for solving
  large scale multiobjective optimization problems,'' in \emph{2013 IEEE
  Congress on Evolutionary Computation}, 2013, pp. 2758--2765.

\bibitem{RN85}
Y.~Tian, X.~Zheng, X.~Zhang, and Y.~Jin, ``Efficient large-scale
  multi-objective optimization based on a competitive swarm optimizer,''
  \emph{IEEE Transactions on Cybernetics}, pp. 1--13, 2019.

\bibitem{RN90}
C.~He, S.~Huang, R.~Cheng, K.~C. Tan, and Y.~Jin, ``Evolutionary multiobjective
  optimization driven by generative adversarial networks (gans),'' \emph{IEEE
  Transactions on Cybernetics}, pp. 1--14, 2020.

\bibitem{10.1145/2480741.2480752}
\BIBentryALTinterwordspacing
M.~\v{C}repin\v{s}ek, S.-H. Liu, and M.~Mernik, ``Exploration and exploitation
  in evolutionary algorithms: A survey,'' \emph{ACM Comput. Surv.}, vol.~45,
  no.~3, jul 2013. [Online]. Available:
  \url{https://doi.org/10.1145/2480741.2480752}
\BIBentrySTDinterwordspacing

\bibitem{10.5555/3295222.3295349}
A.~Vaswani, N.~Shazeer, N.~Parmar, J.~Uszkoreit, L.~Jones, A.~N. Gomez,
  L.~Kaiser, and I.~Polosukhin, ``Attention is all you need,'' in
  \emph{Proceedings of the 31st International Conference on Neural Information
  Processing Systems}, ser. NIPS'17.\hskip 1em plus 0.5em minus 0.4em\relax Red
  Hook, NY, USA: Curran Associates Inc., 2017, p. 6000–6010.

\bibitem{RN82}
X.~Zhang, Y.~Tian, R.~Cheng, and Y.~Jin, ``A decision variable clustering-based
  evolutionary algorithm for large-scale many-objective optimization,''
  \emph{IEEE Transactions on Evolutionary Computation}, vol.~22, no.~1, pp.
  97--112, 2018.

\bibitem{9047876}
Y.~Tian, C.~Lu, X.~Zhang, K.~C. Tan, and Y.~Jin, ``Solving large-scale
  multiobjective optimization problems with sparse optimal solutions via
  unsupervised neural networks,'' \emph{IEEE Transactions on Cybernetics},
  vol.~51, no.~6, pp. 3115--3128, 2021.

\bibitem{YI2020470}
\BIBentryALTinterwordspacing
J.-H. Yi, L.-N. Xing, G.-G. Wang, J.~Dong, A.~V. Vasilakos, A.~H. Alavi, and
  L.~Wang, ``Behavior of crossover operators in nsga-iii for large-scale
  optimization problems,'' \emph{Information Sciences}, vol. 509, pp. 470--487,
  2020. [Online]. Available:
  \url{https://www.sciencedirect.com/science/article/pii/S0020025518308016}
\BIBentrySTDinterwordspacing

\bibitem{RN257}
\BIBentryALTinterwordspacing
H.~Hong, K.~Ye, M.~Jiang, D.~Cao, and K.~C. Tan, ``Solving large-scale
  multiobjective optimization via the probabilistic prediction model,''
  \emph{Memetic Computing}, 2022. [Online]. Available:
  \url{https://doi.org/10.1007/s12293-022-00358-9}
\BIBentrySTDinterwordspacing

\bibitem{RN219}
Z.~Wang, H.~Hong, K.~Ye, M.~Jiang, and K.~C. Tan, ``Manifold interpolation for
  large-scale multi-objective optimization via generative adversarial
  networks,'' \emph{IEEE Transactions on Neural Networks and Learning Systems},
  2021.

\bibitem{DBLP:journals/corr/abs-2012-12556}
\BIBentryALTinterwordspacing
K.~Han, Y.~Wang, H.~Chen, X.~Chen, J.~Guo, Z.~Liu, Y.~Tang, A.~Xiao, C.~Xu,
  Y.~Xu, Z.~Yang, Y.~Zhang, and D.~Tao, ``A survey on visual transformer,''
  \emph{CoRR}, vol. abs/2012.12556, 2020. [Online]. Available:
  \url{https://arxiv.org/abs/2012.12556}
\BIBentrySTDinterwordspacing

\bibitem{RN110}
K.~Deb and H.~Jain, ``An evolutionary many-objective optimization algorithm
  using reference-point-based nondominated sorting approach, part i: Solving
  problems with box constraints,'' \emph{IEEE Transactions on Evolutionary
  Computation}, vol.~18, no.~4, pp. 577--601, 2014.

\bibitem{7155533}
X.~Ma, F.~Liu, Y.~Qi, X.~Wang, L.~Li, L.~Jiao, M.~Yin, and M.~Gong, ``A
  multiobjective evolutionary algorithm based on decision variable analyses for
  multiobjective optimization problems with large-scale variables,'' \emph{IEEE
  Transactions on Evolutionary Computation}, vol.~20, no.~2, pp. 275--298,
  2016.

\bibitem{9082904}
C.~He, S.~Huang, R.~Cheng, K.~C. Tan, and Y.~Jin, ``Evolutionary multiobjective
  optimization driven by generative adversarial networks (gans),'' \emph{IEEE
  Transactions on Cybernetics}, vol.~51, no.~6, pp. 3129--3142, 2021.

\bibitem{RN96}
R.~Cheng, Y.~Jin, M.~Olhofer, and B.~sendhoff, ``Test problems for large-scale
  multiobjective and many-objective optimization,'' \emph{IEEE Transactions on
  Cybernetics}, vol.~47, no.~12, pp. 4108--4121, 2017.

\bibitem{RN92}
Y.~Tian, R.~Cheng, X.~Zhang, and Y.~Jin, ``Platemo: A matlab platform for
  evolutionary multi-objective optimization [educational forum],'' \emph{IEEE
  Computational Intelligence Magazine}, vol.~12, no.~4, pp. 73--87, 2017.

\bibitem{Haynes2013}
\BIBentryALTinterwordspacing
W.~Haynes, \emph{Wilcoxon Rank Sum Test}.\hskip 1em plus 0.5em minus
  0.4em\relax New York, NY: Springer New York, 2013, pp. 2354--2355. [Online].
  Available: \url{https://doi.org/10.1007/978-1-4419-9863-7\_1185}
\BIBentrySTDinterwordspacing

\bibitem{RN98}
E.~Zitzler, L.~Thiele, M.~Laumanns, C.~M. Fonseca, and V.~G.~d. Fonseca,
  ``Performance assessment of multiobjective optimizers: an analysis and
  review,'' \emph{IEEE Transactions on Evolutionary Computation}, vol.~7,
  no.~2, pp. 117--132, 2003.

\bibitem{1583625}
L.~While, P.~Hingston, L.~Barone, and S.~Huband, ``A faster algorithm for
  calculating hypervolume,'' \emph{IEEE Transactions on Evolutionary
  Computation}, vol.~10, no.~1, pp. 29--38, 2006.

\bibitem{jiang2020fast}
M.~Jiang, Z.~Wang, L.~Qiu, S.~Guo, X.~Gao, and K.~Tan, ``A fast dynamic
  evolutionary multiobjective algorithm via manifold transfer learning,''
  \emph{IEEE Transactions on Cybernetics}, 2020.

\bibitem{jiang2020individual}
M.~Jiang, Z.~Wang, S.~Guo, X.~Gao, and K.~Tan, ``Individual-based transfer
  learning for dynamic multiobjective optimization,'' \emph{IEEE Transactions
  on Cybernetics}, 2020.

\end{thebibliography}

\end{document}